\documentclass[conference]{IEEEtran}
\IEEEoverridecommandlockouts
\usepackage{cite}
\usepackage{amsmath,amssymb,amsfonts}
\usepackage{algpseudocode}
\usepackage{hyperref}
\usepackage{graphicx}
\usepackage{textcomp}
\usepackage{pgfplots}
\usepackage{hyperref}
\pgfplotsset{width=8cm,compat=1.9}

\usepackage{xcolor}
\def\BibTeX{{\rm B\kern-.05em{\sc i\kern-.025em b}\kern-.08em
    T\kern-.1667em\lower.7ex\hbox{E}\kern-.125emX}}

\def\footnoterule{\relax%
  \kern-5pt
  \hbox to \columnwidth{\hfill\vrule width 0.5\columnwidth height 0.4pt\hfill}
  \kern4.6pt}
\makeatother

\begin{document}

\title{Bangla Grammatical Error Detection Using T5 Transformer Model}

\author{\IEEEauthorblockN{H.A.Z. Sameen Shahgir}
\IEEEauthorblockA{\textit{Computer Science and Engineering} \\
\textit{Bangladesh University of Engineering and Technology}\\
1805053@ugrad.cse.buet.ac.bd}
\and
\IEEEauthorblockN{Khondker Salman Sayeed}
\IEEEauthorblockA{\textit{Computer Science and Engineering} \\
\textit{Bangladesh University of Engineering and Technology }\\
1805050@ugrad.cse.buet.ac.bd}

}

\maketitle

\begin{abstract}
This paper presents a method for detecting grammatical errors in Bangla using a Text-to-Text Transfer Transformer (T5) Language Model \cite{raffel2020exploringm}, using the small variant of BanglaT5\cite{bhattacharjee2022banglanlg}, fine-tuned on a corpus of 9385 sentences where errors were bracketed by the dedicated symbol \textbf{\$} \cite{bengali-ged}. The T5 model was primarily designed for translation and is not specifically designed for this task, so extensive post-processing was necessary to adapt it to the task of error detection. Our experiments show that the T5 model can achieve low Levenshtein Distance in detecting grammatical errors in Bangla, but post-processing is essential to achieve optimal performance. The final average Levenshtein Distance after post-processing the output of the fine-tuned model was 1.0394 on a test set of 5000 sentences. This paper also presents a detailed analysis of the errors detected by the model and discusses the challenges of adapting a translation model for grammar. Our approach can be extended to other languages, demonstrating the potential of T5 models for detecting grammatical errors in a wide range of languages.
\end{abstract}


\section{Introduction}
In an increasingly digital world, the ability to communicate effectively in written form has become a crucial skill. With the rise of digital communication platforms, such as email, instant messaging, and social media, written communication has become more pervasive than ever before. However, with this increased reliance on written communication comes a new set of challenges, including the need for accurate and effective grammar usage. 

Grammar errors can impede effective communication and have serious consequences, especially in professional and academic settings where clarity and precision are paramount. Grammar errors can also impact the credibility of the writer and create confusion for the reader. In recent years, the development of deep learning models for grammar error detection (GED) and grammar error correction (GEC) has become an increasingly important area of research.

One product of this extensive research is --- Grammarly. It is one of the most ubiquitous grammar correction tools available today, with millions of users around the world. This tool uses the GECToR model \cite{omelianchuk2020gector} for error detection and correction. This model implements a tagging-based approach for error detection using an encoder and then uses a generative approach to correct that error based on the detection using a seq2seq model. This approach achieves state-of-the-art results on canonical GEC evaluation datasets based on F-score results. This makes it a valuable resource for individuals and organizations that rely on written communication. However, it is important to note that Grammarly and other similar tools are currently only available for a limited number of languages, primarily English.

Some research work has been done in GED and GEC in Bangla \cite{noshin2020bangla} \cite{hossain2021development} but to the best of our knowledge, no work leveraging transformer models has yet been done in Bangla. As mentioned before, GEC in English has already reached a commercially viable stage and notable progress has been achieved using both seq2seq \cite{grundkiewicz2019neural} \cite{kiyono2019empirical} and BERT-based models \cite{omelianchuk2020gector}. Both deliver comparable performance \cite{omelianchuk2020gector} but the seq2seq models are easier to train albeit with much slower inference. We ultimately decided on using the T5 model \cite{raffel2020exploring}, pre-trained on a large Bangla corpus \cite{bhattacharjee2022banglanlg}. We tested both the base (220M parameters) and the small (60M parameters) variants of BanglaT5 and found the smaller model to perform slightly better within our computing budget.

T5 or Text-to-Text Transfer Transformer \cite{raffel2020exploring}, is a Transformer based architecture that uses a text-to-text approach. It adds a causal decoder to the bidirectional architecture of BERT \cite{devlin2018bert}. The difference with the basic encoder-decoder transformer architecture \cite{vaswani2017attention} is that t5 uses relative positional embedding and layer norm at the start of each block and the end of the last block. Other than that, t5 and the basic encoder-decoder transformers are the same in architecture. T5 was trained with the goal of unifying all NLP tasks with a single text-to-text model. By that goal, banglat5 \cite{bhattacharjee2022banglanlg} was trained on a massive Bengali pretraining corpus Bangla2B+ \cite{bhattacharjee2021banglabert}, sized 27.5GB. This allows banglat5 to achieve state-of-the-art results on most Bengali text generation tasks. Therefore, leveraging transfer learning from an enormous Bengali text corpus --- banglat5 is an ideal candidate to consider for Bengali GED and GEC tasks.  

\begin{figure*}[h]
    \centering
    \includegraphics[width=0.8\columnwidth]{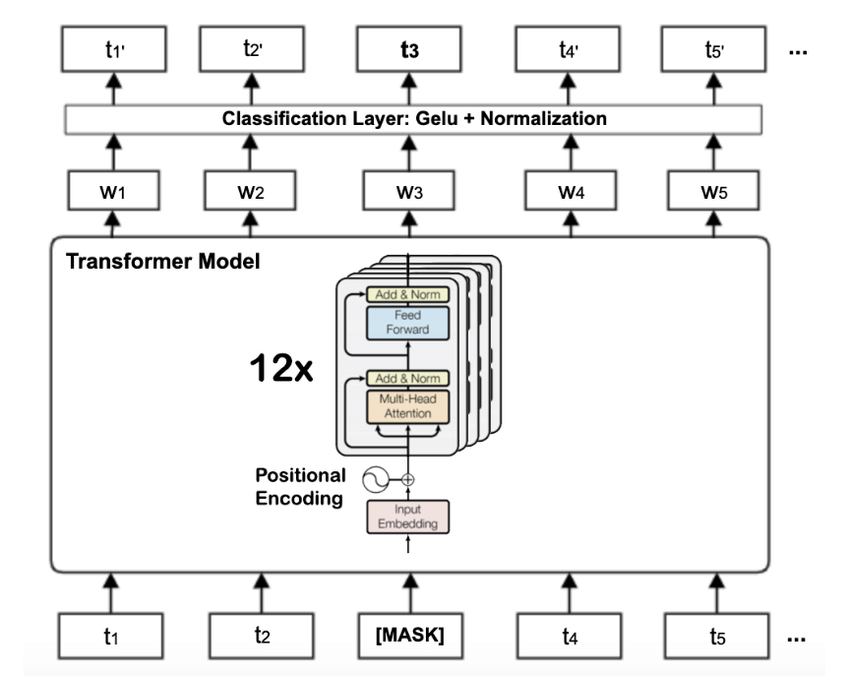}
    \includegraphics[width=1.2\columnwidth]{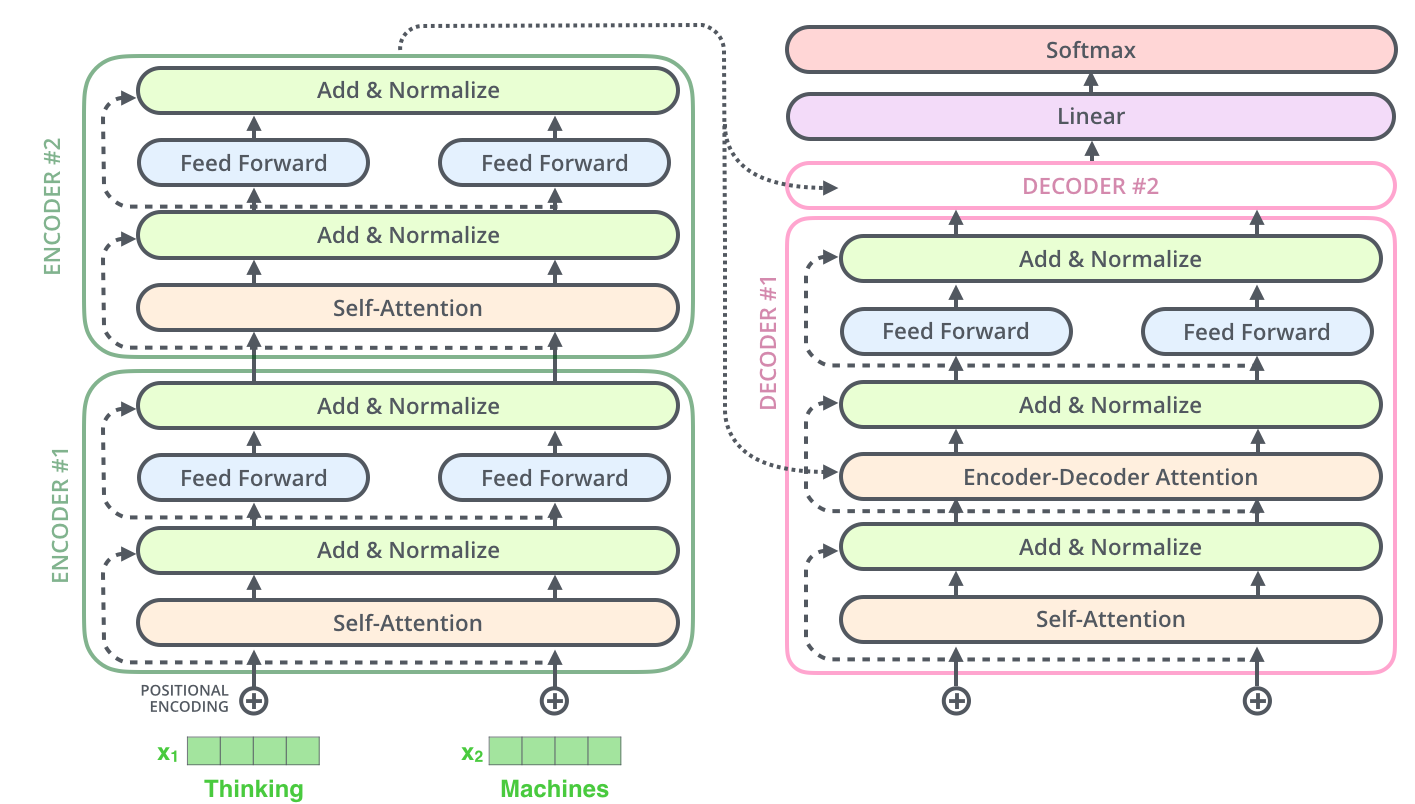}
    \caption{Encoder based BERT architecture (left) vs Encoder-Decoder based text-to-text transformer architecture (right) (source: \href{https://jalammar.github.io/illustrated-transformer/}{https://jalammar.github.io/illustrated-transformer/})}
    \label{fig:1}
\end{figure*}

\section{Methodology}
\subsection{Model Selection}
Currently, BERT (Bidirectional Encoder Representations from Transformers) and its variants are the best-performing models on tasks such as token classification\cite{malmasi2022semeval} and sentence classification\cite{patwa2020semeval}. BanglaBERT reproduced this finding when trained specifically on Bangla corpus\cite{bhattacharjee2021banglabert}. Although GED can be formulated as either a token classification problem or a sentence classification problem, both possess several challenges. When presented as a token classification task, punctuation becomes a particular issue since most punctuations represent a pause and are hard to distinguish. Another challenge is tokens which are missing altogether. It can be hypothesized that BERT does have the ability to detect the logical inconsistency in a sentence that arises from missing tokens due to its deep encoder architecture but marking the position of missing tokens is a challenge. On the other hand, when posed as a sentence classification problem, we find that BERT can classify sentences are either error-free or with errors well but cannot mark the erroneous section itself. Recently, sequence-to-sequence (seq2seq) models such as the T5 \cite{JMLR:v21:20-074} have achieved state-of-the-art performance on standard Grammatical Error Correction (GEC) benchmarks\cite{bryant2019bea}. Such models \cite{grundkiewicz2019neural} \cite{kiyono2019empirical} have been trained specifically on synthetic GEC datasets (as opposed to general translation datasets). But since the model must generate the entire output sequence, including the parts which were correct to begin with, inference is slow. The BERT-based GECToR \cite{omelianchuk2020gector} presents another way for GEC - a token classification approach where errors are mapped to the 5000 error-correcting transformations (one for each token in vocabulary and some token independent transformations) which correct the errors algorithmically. The resulting model is up to 10 times faster than comparable seq2seq models but as before, this requires a synthetic pretraining corpus.

For Bangla Grammar Error Detection we decided on the small variant of BanglaT5\cite{bhattacharjee2022banglanlg} with 60M parameters. The smaller model allowed for larger batch sizes, faster experimentation and hyper-parameter tuning when compared to the standard BanglaT5 model with 220M parameters while delivering similar performance on our training set (9385 pairs). Experimentation on the larger T5 models using the full available dataset (19385 pairs) and evaluating a BERT-based approach similar to GECToR is left for future work.

\subsection{Dataset Analysis}
The training set consisted of 19385 sentence pairs in total, containing both error-free sentences and sentences with errors. The major error types are:
The major error types are:
\begin{enumerate}
\item Single word error
\item Multi-word error
\item Wrong Punctuation 
\item Punctuation Omission
\item Merge Error
\item Form/Inflection Error
\item Unwanted space error
\item Hybrid

\end{enumerate}
The errors are each bracketed by a designated symbol \textbf{\$} and are not differentiated from each other.

\begin{table}[h!]

    \centering
    
    \caption{Dataset characteristics}
        \begin{tabular}{|l|r|r|r|}
        \hline
        \textbf{Split} & \textbf{Total} & \textbf{With Error} &  \textbf{Num. Errors} \\
        \hline
        DataSetFold1 & 9385 & 3693 & 7133\\
        DataSetFold2 & 10000 & 4393 & 7352 \\
        Test & 5000 & - & - \\
        \hline
        \end{tabular}
        
    \label{tab:my_label}
\end{table}

We used DataSetFold1 for the fine-tuning of the T5 model and both DataSetFold1 and DataSetFold2 for the crucial post-processing steps.

\subsection{External Dataset}
We collected a word list of 311 archaic Bangla verb words which were consistently marked as errors in the training dataset. We collected said word list with the aid of \href{https://github.com/features/copilot}{Github Copilot}. This data was used in our regular expression based approach to GED.

\subsection{Pre-processing}
Not wanting to shift the distribution of the train set from the test set, we kept pre-processing to a minimum. The sentences were normalized and tokenized using the normalizer and tokenizer used in pretraining \cite{bhattacharjee2022banglanlg}. One notable point is that we omitted newline characters when present inside sentences since it interfere with the way the T5 model reads in sentences.

\subsection{Training}
Through experimentation on an 80-20 split of DataSetFold1 between training and validation set and using an effective batch size of 128, we determined 120 epochs to be a good stopping point before the model starts to over-fit. Then we used the entirety of DataSetFold1 for 120 epochs of training. Since the task was to predict 5000 test sentences while training only on 9385 training pairs, we determined that keeping any significant segment for validation and early stopping would be detrimental to overall performance.

\begin{figure}[h!]
    \centering
    \begin{tikzpicture} \label{plt:1}
        \begin{axis}[
            xlabel={Epoch},
            ylabel={Training Loss},
            xmin=0, xmax=130,
            ymin=1.30, ymax=2.2,
            xtick={10, 20, 30, 40, 50, 60, 70, 80, 90, 100, 110, 120},
            ytick={1.30, 1.40, 1.50, 1.60, 1.70, 1.80, 1.90, 2.0, 2.1, 2.2},
            ymajorgrids=true,
            grid style=dashed,
        ]
        
        \addplot[color=black, smooth, thick] table[x index=0, y index=1, col sep=comma, skip first n=11] {training_data.csv};
        \end{axis}
    \end{tikzpicture}
    \caption{Training Loss vs Epoch} \label{fig:1}
\end{figure}
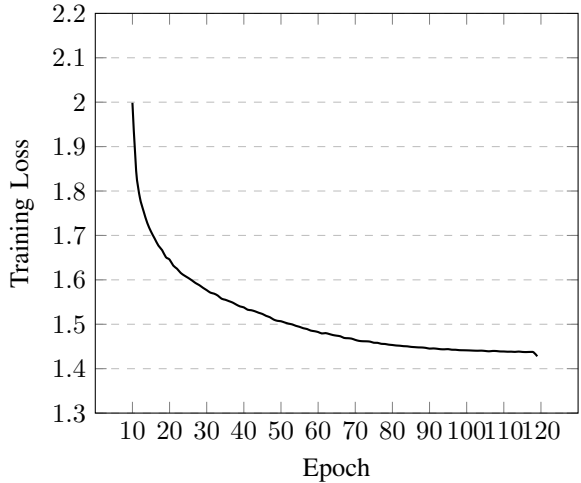

A naive attempt to train the model on the combined DataSetFold1 and DataSetFold2 dataset did not improve our score on the test set. This is likely because introducing new data requires re-tuning the model hyper-parameters. For now, we leave this as future work.

\subsection{Post-processing}
The T5 model was built on the paradigm of unifying all NLP tasks under text-to-text classification and on that front, T5 achieves state-of-the-art results on many GLUE tasks. However, this paradigm does have its shortcoming. Of particular importance in the task of GED when judged by Levensthein Distance is the tendency of T5 models to spell words differently or sometimes change entire words with a close synonym because reproducing the input sequence exactly is as important as marking the errors. This is a particular problem in Bangla GED since the language is still evolving and multiple spellings of the same word are in use concurrently. Furthermore, there exist several unicode characters representing the same Bangla alphabet or symbol, further complicating the reconciliation of the T5 output sequence with its input sequence.

To transform the raw T5 output to a form as close as possible to the input sequence, we present two algorithms. As an optional post-processing of all outputs, we present a third algorithm that does simple error word detection that the model might have missed.

The first is for respelling and correcting the T5 output by comparing it character by character with the input sequence. Beginning with an empty string as corrected output, if the next character is a \textbf{\$} symbol, it is appended to the corrected output string. If the next character of the input and output sentence match, it is appended. If they do not match, the next character of the output string is looked up in a table and if present, the value from the lookup table represents the correction and is appended. This lookup table has been constructed manually by observing common t5 errors. Constructing the table automatically is left for future work.

If character-level corrections fail, the algorithm attempts to make word-level corrections by replacing entire words in the t5 output and then character-level correction is attempted again.

The second algorithm is a regular expression-based approach to GED in case the first algorithm fails to correct the T5 output. Certain common errors are learned from the training dataset and identified in the test set using sub-string replacements.

These two algorithms work in tandem to correct the t5 output. However, should a test sentence already be present in the training dataset, then the error-marked sentence is directly pulled from the training dataset using another lookup table. In a real-world scenario, having a lookup table of the most commonly mistaken sentences or phrases can significantly speed up GED since the need for a large deep learning model is bypassed entirely.

The pseudo-code for the algorithms is in the Appendix.
\vspace{250pt}
\pagebreak
\section{Results and Discussion}
Training the banglat5 small model with 60M parameters on 9385 sentence pairs for 120 epochs with a batch size of 128 and learning rate of $5\times10^{-4}$ with AdamW Optimizer and a linear learning rate scheduler yielded a final Levensthein Score of 1.0394 on 5000 test sentences. The effect of the multiple post-processing steps is presented below, serving as a short ablation study of our methodology. Average Levenshtein distance data on the test dataset was collected from submissions to \href{https://www.kaggle.com/competitions/bengali-ged/overview}{EEE DAY 2023 Datathon}. The private and public scores are based on a 50-50 split of the 5000 test sentences. We calculated the total Aggregated distance by averaging the two.

\begin{table}[h]
\centering
\caption{ \textbf{Average Levenshtein Score}
\\
CC = Character Correction \\
WC = Word Correction \\
R = Regex Model\\
L = Lookup Table \\
P2 = Post Processing 2 \\
}

\begin{tabular}{|l|c|c|c|c|c|c|c|}
\hline
Model & Regex & Match & Private & Public & Aggregated \\
\hline
Raw & - & - & 3.216 & 3.208 & 3.212 \\
No Corr. & - & - & 1.5072 & 1.5168 & 1.512 \\
R & 5000 & - & 1.1896 & 1.1916 & 1.1906 \\
CC & - & - & 1.1072 & 1.134 & 1.1206 \\
CC+R & 107 & - & 1.0732 & 1.1048 & 1.089 \\
CC+WC+R & 42 & - & 1.072 & 1.1012 & 1.0866 \\
\textbf{CC+WC+R+L} & \textbf{40} & \textbf{253} & \textbf{1.0224} & \textbf{1.0588} & \textbf{1.0416} \\
\textbf{CC+WC+R+L+P2} & \textbf{40} & \textbf{253} & \textbf{1.0224} & \textbf{1.0564} & \textbf{1.0394} \\
\hline
\end{tabular}
\label{table:2}
\end{table}

After character-level corrections, the T5 output still had a severe mismatch with the original input in 107 sentences. These arise mainly from mainly two causes, entire words replaced or sentences that exceed the maximum input token limit (256) of the model. Using only the regex-based algorithm yields a modest score of 1.1906. But using it to handle 107 sentences that couldn't be corrected resulted in a significant improvement (1.0866). Finally, the lookup table also modestly improves the Levensthein score (1.0394) by looking up 253 sentences with exact matches in the training dataset.

\section{Conclusion and Future Work}
In conclusion, we trained a T5 model for Grammatical Error Detection and evaluated its performance using Levenshtein distance. Although it's difficult to compare our results to previous work that typically uses the F1 score, our model achieved good performance on the dataset we used. However, we acknowledge that we only used 50\% of the dataset and the entire dataset may have improved our results. Additionally, using T5 base instead of T5 small may have improved our performance with hyperparameter tuning.

We also noted that preprocessing could have rooted out spelling errors, leaving more difficult semantic errors for the T5 model to handle. Moreover, we identified that the post-processing step could be automated to improve the performance further.

Looking forward, we suggest exploring a BERT-based approach like GECToR\cite{omelianchuk2020gector} for Grammatical Error Detection. Overall, our work demonstrates the potential of T5 models for Grammatical Error Detection and provides a foundation for future work in this field.

\bibliography{refs}

\pagebreak
\section{Appendix}
\subsection{Pseudocode for Correcting Mistatches in T5 Ouput}
\begin{verbatim}
function t5_output_correction(t5_output, t5_input):
    corrected_output = ""
    i1 = 0
    i2 = 0
    attempt_word_corr = True
    while True:
        if i1 == len(t5_input) and i2 == len(t5_output):
            return corrected_output
        
        if t5_input[i1] == t5_output[i2]:
            corrected_output += t5_input[i1]
            i1 += 1
            i2 += 1
            continue
            
        if t5_output[i2] == "$":
            corrected_output += "$"
            i2 += 1
            continue
            
        if t5_output[i2] in character_lookup.keys():
            if t5_input[i1] == character_lookup[t5_output[i2]]:
                corrected_output += t5_input[i1]
                i1 += 1
                i2 += 1
                continue
                
        if attempt_word_corr == True:
            t5_output = word_correction(t5_output, t5_input)
            i1 = 0
            i2 = 0
            attempt_word_corr = False
            continue
            
        return regex_correction(t5_input)

        
\end{verbatim}

\subsection{Pseudocode for Regex-based Error Detection}
\begin{verbatim}
function regex_correction(sentence):
    for word in common_errors_set:
        sentence = sentence.replace(word, "$"+word+"$")

    for rule in regex_rules:
        sentence = sentence.replace(rule, "$"+rule"$")
\end{verbatim}
\end{document}